# Streamlining Social Media Information Retrieval for COVID-19 Research with Deep Learning

Yining Hua, MS[1,2,3,1]; Jiageng Wu, MS[4]; Shixu Lin, BS[4]; Minghui Li, BS [4]; Yujie Zhang, BS[4]; Dinah Foer, MD[3]; Siwen Wang, MD[1]; Peilin Zhou, MS[5]; Jie Yang, PhD[6, b#]; Li Zhou, MD, PhD[3, c#]

[1] Department of Epidemiology, Harvard Chan School of Public Health, Boston, MA, USA
[2] Department of Biomedical Informatics, Harvard Medical School, Boston, MA, USA
[3] Division of General Internal Medicine and Primary Care, Department of Medicine, Brigham and Women's Hospital and Harvard Medical School, Boston, MA, USA
[4] School of Public Health, Zhejiang University School of Medicine, Hangzhou, Zhejiang, China
[5] Thrust of Data Science and Analytics, The Hong Kong University of Science and Technology (Guangzhou), Guangzhou, Guangdong, China.
[6] Division of Pharmacoepidemiology and Pharmacoeconomics, Department of Medicine, Brigham and Women's Hospital, Harvard Medical School, Boston, MA, USA

## ABSTRACT

**Objective**: Social media-based public health research is crucial for epidemic surveillance, but most studies identify relevant corpora with keyword-matching. This study develops a system to streamline the process of curating colloquial medical dictionaries. We demonstrate the pipeline by curating a UMLS-colloquial symptom dictionary from COVID-19-related tweets as proof of concept.

**Methods**: COVID-19-related tweets from February 1, 2020, to April 30, 2022 were used. The pipeline includes three modules: a named entity recognition module to detect symptoms in tweets; an entity normalization module to aggregate detected entities; and a mapping module that iteratively maps entities to Unified Medical Language System concepts. A random 500 entity sample were drawn from the final dictionary for accuracy validation. Additionally, we conducted a symptom frequency distribution analysis to compare our dictionary to a pre-defined lexicon from previous research.

**Results:** We identified 498,480 unique symptom entity expressions from the tweets. Pre-processing reduces the number to 18,226. The final dictionary contains 38,175 unique expressions of symptoms that can be mapped to 966 UMLS concepts (accuracy = 95%). Symptom distribution analysis found that our dictionary detects more symptoms and is effective at identifying psychiatric disorders like anxiety and depression, often missed by pre-defined lexicons.

**Conclusions:**

This study advances public health research by implementing a novel, systematic pipeline for curating symptom lexicons from social media data. The final lexicon's high accuracy, validated by medical professionals, underscores the potential of this methodology to reliably interpret and categorize vast amounts of unstructured social media data into actionable medical insights across diverse linguistic and regional landscapes.

**Keywords**: Social Media, Public Health, COVID-19, Named Entity Recognition, Entity Name Normalization, Deep Learning

---
[1] yininghua@g.harvard.edu; [b]jyang66@bwh.harvard.edu; [b]lzhou@bwh.harvard.edu
[#] Shared senior authorship and correspondence.

# 1. BACKGROUND

The efficacy of public health systems faces multiple levels of challenge, from disease recognition to system-based implementation.[1] This was notably evident in the United States during the COVID-19 pandemic, where the varied responses of individual states were often conflicting, complicating public understanding, mandate compliance, and resource allocation.[2,3] Downstream health outcomes were also differentially impacted. Such variability underscores the need for a robust public health surveillance system that can offer tailored, region-specific guidance by tracking and synthesizing variable data. The pandemic also revealed a critical concern: the potential for large-scale health crises to overwhelm hospitals, further hindering routine care for acute and chronic conditions.[4] These concerns underscored the importance of remote health monitoring and the need for innovative solutions in public health management.

Social media data have shown promising potential for real-time surveillance and large-scale tracking of public reactions to the pandemic. Platforms like Twitter,[2] Reddit, and Facebook, with their ease of access and real-time nature, are extensively used to study recent epidemics.[5] By analyzing vast amounts of individual behavior data, researchers can identify collective and individual behavior patterns.[6,7] However, a key challenge in utilizing social media to study public health discussions is to ensure that the raw data retrieval methods capture a sufficiently representative study sample. As an important task of natural language processing (NLP), information retrieval involves efficiently locating data that meets specific criteria from large repositories, the challenges of which is further amplified by the diverse and informal nature of social media communication.

To retrieve relevant information, most social media-based public health studies rely on rule-based algorithms[8] that typically involve matching the corpus with a predefined lexicon. However, medical keyword-based searches in social media are inherently susceptible to the incompleteness of the lexicon for several reasons: (1) Social media users often use personalized colloquialisms rather than formal clinical or medical terms to describe their symptoms; (2) it is hard to establish a comprehensive lexicon to cover the colloquial descriptions of a broad spectrum of symptoms, and the pre-defined lexicon may also exist systematic bias or lag; (3) the difficulty of mapping the self-reported symptoms into clinically-defined symptoms make the social-media findings hard to reconcile with clinical study findings.

Although several works intend to introduce deep learning-based models to identify potential research corpora,[3,9] such methods rely on a large-scale labeled dataset and are hard to scale with the evolution of disease symptoms. The inefficient re-label and re-train approach cannot provide clinicians and policymakers with timely information. In addition, deep learning-based models are time-consuming and computationally intensive, making it difficult to process large-scale social media data quickly. The intricate nature of deployments and the complexity of configuring environments pose significant obstacles to the adoption of artificial intelligence (AI) and computational tools by researchers outside the computer science domain.

---

[2] Renamed to X in July 2023.

The goal of this study is to standardize and simplify the process of creating predefined keywords for more representative information retrieval in social media-based COVID-19 studies. To do so, we aimed to develop and propose a framework for creating a dictionary that maps Unified Medical Language System (UMLS)[10] concepts to colloquial medical vocabulary from social media corpora. In turn, this framework can be scaled and generalized to support future public health research using social media. We divide this strategy into three steps: large-scale medical entity extraction, entity normalization, and lexicon alignment. All three modules are supported by state-of-the-art natural language processing (NLP) techniques. To demonstrate the framework, we conducted a proof-of-concept study in which we applied the framework to a large Twitter dataset to build a symptom dictionary and quantitatively analyzed the results.

## 2. SIGNIFICANCE

This study contributes to the computational public health field in the following three ways:

1. It proposes a pipeline framework for standardizing the development of predefined dictionaries for information retrieval for COVID-19 studies using social media.
2. Using this framework, we developed and released a comprehensive dictionary of UMLS concepts and symptom pairs derived from a massive volume of COVID-19-related tweets. This dictionary is accessible online at https://github.com/ningkko/UMLS_colloquial.
3. By matching symptoms using the constructed colloquial dictionary and comparing them with symptom frequencies matched without using the colloquial dictionary, we analyzed the efficiency of this dictionary.

## 3. MATERIALS AND METHODS

### 3.1. Data Sources

This research has been approved by the Institutional Review Board of Zhejiang University and Mass General Brigham. Following the automation rules and data security policy of Twitter (now known as X), we used Tweet unique identifiers (Tweet ID) provided in an open-source COVID-19 tweet dataset[11] to download related tweets from February 1, 2020, to April 30, 2022 (93 weeks) through the Twitter Application Programming Interface (Twitter API). We focused on English tweets; non-English tweets were removed at the tweet downloading stage. In addition, our preliminary sampling study found that retweets and tweets with Uniform Resource Locators (URLs) generally do not contain user statements. Since this study primarily focuses on the terms used by the public, we also removed such tweets. 18,220,993 tweets remained included in the raw dataset of COVID-19-related tweets for study analysis.

## 3.2. Overall strategy

**Figure 1** shows the overall workflow of the proposed strategy in creating a colloquial lexicon for symptoms using COVID-19-related discussion on Twitter, as well as an example. The strategy comprises a named entity extraction module that extracts as many relevant entities as possible from massive candidate corpora. First, we trained a Bidirectional Encoder Representations from Transformers (BERT)[12] model to extract symptom entities from tweets. Second, we preprocessed the extracted entities to reduce the number of symptoms for normalization. Last, we mapped the preprocessed entities to UMLS concepts by iteratively training and applying a deep learning model.

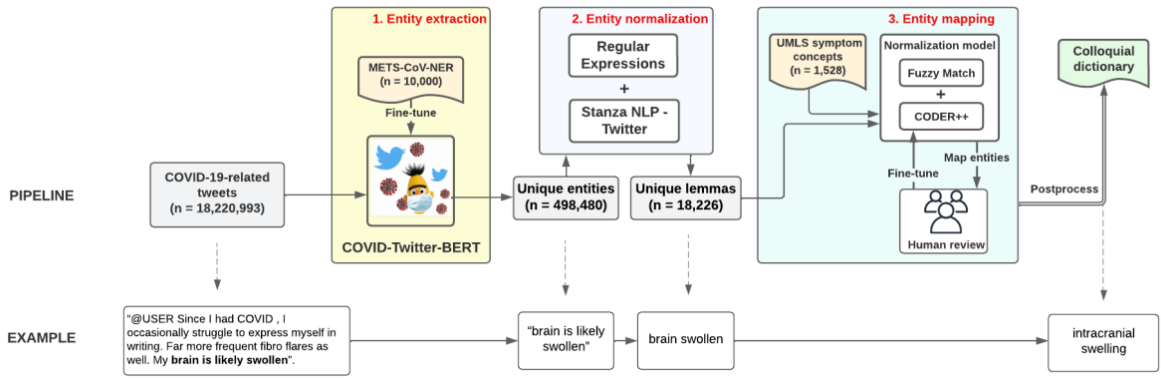

**Figure 1.** Overview of the colloquial lexicon curation strategy creating a UMLS-colloquial symptom dictionary using COVID-19-related tweets. Sensitive data such as usernames are removed in the example.

## 3.3. Entity Extraction

The first module of our strategy is an NER model that extracts entities of interest from the entire corpus of tweets. We used a social media medical NER model developed and validated in our previous work,[13] which has been proven to be the state-of-the-art model for this task. This model was developed based on COVID-Twitter-BERT (CT-BERT)[14] and was fine-tuned on METS-CoV,[13] a dataset homogeneous to our dataset. METS-CoV comprises 10,000 COVID-19-related tweets with 19,072 entities of four medical entity types (Disease, Drug, Symptom, and Vaccine) and three general entity types (Person, Location, and Organization). The NER model had a mean F1 score of 81.85% (standard deviation: 0.53%) with five-fold cross-validation on the test set of METS-CoV. More information including a detailed description of the model structure is presented in Appendix A.

After fine-tuning the NER model, we fed all candidate tweets to the module to extract symptom entities for each tweet. Before extraction, we followed the pre-cleaning steps previously described[13] to remove HTML tags with the Texthero package[3] in Python and replace Unicode emotion icons with textual ASCII representations using the emoji package in Python. Tweets with no identified symptoms in this step were removed.

### 3.4. Entity normalization

This module serves the purpose of minimizing the manual review workload required for the subsequent entity mapping module. It consists of two primary steps: 1) pre-cleaning the extracted entities using regular expressions to deduplicate them, and 2) utilizing the Twitter package of Stanza (Twitter-Stanza)[15] for lemmatization.

In the initial step, regular expressions are employed to eliminate stopwords and replace specific patterns we summarized from the entities with more generalized forms. This pre-cleaning process helps reduce the computational time required for lemmatization. More specifically, phrase listed in Appendix B are systematically replaced by replacement patterns. Additionally, any content within parentheses and the parentheses themselves are removed. Furthermore, all punctuation marks, except commas and asterisks, are eliminated. We observed that commas are valuable in distinguishing multiple symptoms and compound symptoms, while asterisks are frequently used to mask letters.

After the pre-cleaning step, Twitter-Stanza is utilized for lemmatizing the entities. However, since Twitter-Stanza unifies numeric values to "number," the keyword "number" is removed from the lemmatized entities. Subsequently, the pre-cleaning rules are applied again to merge the lemmatized entities further.

To remove trivial expressions, we eliminate lemmatized entities (along with their corresponding symptom entities) that occurred <10 times during the study period. This step helps refine the entity selection by filtering out entities that appear infrequently and may not contribute significantly to the analysis.

### 3.5. Entity clustering

#### 3.5.1. The module structure

The entity clustering module iteratively maps colloquial expressions to unique concepts. The UMLS ontology framework was chosen as a reference in our case because it is a widely used ontology framework in biomedical sciences, providing hierarchical normalization structures among controlled vocabularies.

The mapping module, illustrated in Figure 2, employs an ensemble approach to assign each precleaned entity the most likely corresponding UMLS label. This ensemble approach consists of a semantic part and a lexical part. In the semantic part, CODER++[16] is utilized to embed each entity and UMLS label. CODER++ is a BERT-based framework specifically designed for medical term normalization and demonstrates superior performance compared to other widely used BERT models. It uses contrastive learning with medical terms and calculates similarities from term-relationship-term triplets derived from the UMLS knowledge graph. The cosine similarity metric is utilized to find the most similar UMLS label for each entity.

---

[3] https://texthero.org/

The dictionary part of the iterative mapping module involved using fuzzy matching with the Levenshtein distance to calculate the similarity between each entity-UMLS label pair. The Levenshtein distance is a measure of the minimum number of single-character edits (insertions, deletions, or substitutions) required to transform one string into another. For each lemma, only the concept with the highest similarity score was retained. In cases where fuzzy matching and CODER++ produced different results, both were kept for manual validation. A threshold value ($\tau$) was introduced to remove entities with low similarity scores. The values of $\tau$ were set to 0.8 as determined by identifying the elbow points in the first round of normalization. More details on determining $\tau$ is shown in Appendix C.

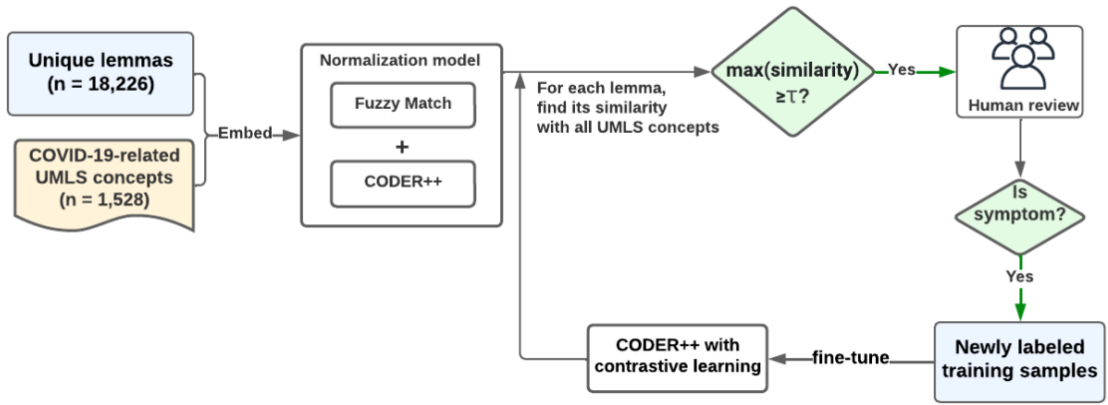

**Figure 2.** A detailed illustration of the iterative normalization system.

### 3.5.2. *Experiment settings*

#### 3.5.2.1 *Reducing Candidate UMLS Concepts*

Since UMLS encompasses over two million names for approximately 900,000 concepts, a subset of UMLS concepts from the PASCLex post-COVID lexicon[17] was selected to reduce computation cost for mapping. PASCLex is a comprehensive lexicon developed based on extensive hospital clinical notes from SARS-CoV-2-positive patients during their post-acute infection period at the Massachusetts General Brigham health system (United States). PASCLex contains 5,377 synonyms derived from 1,528 UMLS terms, ensuring the comprehensiveness and relevance of the lexicon to post-COVID symptoms.

*3.5.2.2 Exit Condition of the Training Loop*

After each round of model prediction, a random sample of 50 pairs of lemmas and predicted UMLS concepts were manually checked for validation. If the model accuracy dropped below 10%, the iteration and the rest of the unaligned entities were abandoned, and the iterative training stopped. If not, we manually examine all the predicted pairs to create new training data for the next iteration and add the manually examined pairs to the final dictionary.

*3.5.2.3 Manual Validation*

In each round of manual validation, three medical students assigned each entity a label from {0, 1, 2}, where 0 represents incorrect predictions, 1 represents correct predictions, and 2 represents non-symptom entities. Specifically, the symptom-label pairs were divided into roughly three equal sets for each iteration, with each medical student responsible for reviewing two sets. After each iteration, Cohen's Kappa was calculated to assess the level of agreement among the annotations. Any disagreements were resolved through discussions between the two annotators assigned to the set. Entities with a label of 2 (non-symptom entities) were removed. Entities with labels 0 and 1 were reorganized as training samples for fine-tuning CODER++ with contrastive learning.

In the preliminary validation process, we found that colloquial terms often had ambiguous meanings. Therefore, an additional validation step in which previously unseen pairs were assigned to each annotator was implemented to enhance the quality of the final dictionary. For this phase, ten sample tweets were extracted for each lemma, allowing annotators to use context for clarification whenever a lemma appeared ambiguous. The resulting dictionary was considered the finalized version.

As a final step, an attending-level, internal medicine-trained physician (D.F.) conducted a sanity check by reviewing 100 randomly selected lemma-UMLS concept pairs. Each pair was accompanied by ten tweet examples sourced from the final dictionary, to ensure the overall quality and accuracy of the compilation.

*3.5.2.4 Fine-tuning the mapping model*

We employed the SentenceBERT[18] framework, which is optimized for rapid contrastive learning, to fine-tune CODER++. Utilizing Siamese and triplet network architectures, SentenceBERT generates semantically rich sentence embeddings that can be compared efficiently using cosine similarity measures. This approach accelerates the process of identifying the most similar pairs while maintaining a high level of accuracy.

Training data was constructed using hard negative sampling to provide more information for CODER++ to learn. For each UMLS concept $u_i$ in the COVID-19 symptom subset $U$, the correctly matched entities were defined as P($u_i$), and the wrongly matched entities were defined as N($u_i$). If both P($u_i$) and N($u_i$) were non-empty sets, they were used as the training data for ui. If P($u_i$) was empty, ui was considered a positive sample, i.e., P($u_i$) = {$u_i$}. If N($u_i$) was empty, negative samples for ui were randomly sampled from P($u_j$), where $u_j \in U$ and $j \neq i$. Figure 3 shows an example of the training tuples.

We set the maximum length of tokens to 512. The training process was conducted over three epochs with a batch size of 256 via the SentenceTransformer library with Triplet loss. Training took place on an NVIDIA Quadro P6000 GPU with 24 GB of memory, taking around 27 minutes.

**Query:** 'abdominal cramps'

**Positive samples:** 'abdominal cramping', 'abdominal cramp', 'ab cramp'

**Negative samples:** 'abdomen hurt', 'abdominal trauma', 'belly full'

**Figure 3.** An example training tuple generated from hard negative sampling.

### 3.6. Symptom frequency Analysis

We conducted a proof-of-concept study to demonstrate the proposed framework. Specifically, we used the constructed dictionary to extract symptomatic expressions from a collection of tweets related to COVID-19. We downloaded 471,553,966 COVID–19–related tweets from February 1, 2020, to April 30, 2022. These same tweets had been previously studied for drug and symptom networks and symptoms had been identified using a predefined lexicon.[19] Using our framework and lexicon, we then identified tweets with self-reported COVID-19 symptoms. We calculated the occurrences of symptoms identified by our approach and compared our results to the previously published symptom frequencies. In addition, we compared our findings with the symptoms reported by the World Health Organization (WHO)[20] to evaluate if the symptoms aligned with clinically relevant patterns.

## 4. RESULTS

### 4.1. Entity Extraction and Normalization

Applying the NER model to all candidate tweets, we identified 603,3184 symptom entities (498,480 unique) from 4,401,304 tweets. After normalization, 18,226 unique lemmatized symptoms were left for mapping.

We summarized the frequency statistics of all lemmatized entities and lemmatized entities with ≥ 10 occurrences in Table 1. The data indicates a disproportionate focus on a subset of symptoms with a mean of 85 occurrences for all lemmatized entities, but a significantly higher 439 occurrences for entities documented at least 10 times. This focus is further emphasized by the large standard deviations – 2048 for all entities and 4736 for those with at least 10 occurrences—signaling a wide-ranging, uneven distribution of symptom mentions. The interquartile ranges, spanning 1 to 13 for all lemmatized entities and 28 to 114 for those with ≥ 10 occurrences, provide a focused lens on this disparity, underscoring a long-tail distribution of symptoms where a small set of symptoms is heavily discussed, overshadowing a large set of infrequently mentioned symptoms.

**Table 1.** Summary statistics of precleaned symptom entity frequencies in a COVID-19 tweet dataset.

|  | All lemmatized entities | ≥ 10 occurrences |
|---|---|---|
| **Mean** | 85 | 439 |
| **Standard deviation** | 2048 | 4736 |
| **Minimum** | 1 | 10 |
| **Median [Interquartile Range]** | 3 [1,13] | 47 [28, 114] |
| **Maximum** | 268183 | 268183 |

## 4.2. Entity Mapping

Table 2 reports summary statistics of the iterative normalization and annotation process. In the first round of normalization, 6772 entities (37.16%) out of 18,226 were correctly mapped by at least one of the normalization approaches (fuzzy match and CODER++). In the second round, 2133 (27.62%) out of 9373 entity lemmas were mapped correctly. In the third round, only 16% (8 entities) out of 50 sampled entities were correct. Therefore, the normalization process ended with the second iteration. The final lexicon includes 3453 unique lemmas that can be mapped to 646 unique UMLS concepts and 11,625 unique unprocessed symptom entities. The first round of manual validation yielded a Cohen's Kappa of 0.79, indicating substantial agreement. The second round of manual validation saw an improvement in agreement, with a Cohen's Kappa of 0.86, indicating almost perfect agreement. The final physician sanity check identified an accuracy of 95%.

**Table 2**. Summary statistics of the iterative normalization and annotation process.

|  | First round | Second round | Validation |
|---|---|---|---|
| # entities to cluster | 18,226 | 9373 | - |
| # total mapping pairs | 23252 | 9874 | - |
| # entities removed by $\tau = 0.5$ | 3,649 | 1970 | - |
| # remaining mapping pairs to annotate | 19603 | 7403 | 9045 |
| # entities predicted wrong (label 0) | 9373 | 5094 | 5592 |
| # entities predicted correct (label 1) | 6772 | 2273 | 3453 |
| # non-symptom entities (label 2) | 3077 | 537 | 0 |
| **Cohen's Kappa** | 0.79 | 0.86 | - |
| **Prediction accuracy** | 37.16% | 24.25% | - |

Note: The total number of unique pairs is greater than the number of entities to cluster because the lexical and semantic parts may generate different mappings for the same entity

### 4.3. Symptom Distribution Analysis

We identified 2,761,059 tweets with self-reported COVID-19 symptoms, nearly three times as many as in previous studies on the same dataset. We selected 245 UMLS with occurrences greater than or equal to 500 and merged similar UMLS concepts into specific symptoms aligned with the WHO report's symptom scheme for better comparison. Table 3 presents a comparative analysis of physical symptoms derived from three distinct sources: (1) symptoms extracted using our dictionary from self-reporting positive patients on Twitter, (2) symptoms extracted using pre-defined keywords as documented in our prior study utilizing the same dataset, and (3) symptoms reported in the WHO report, primarily based on the Chinese in-hospital population in the early months of the COVID-19 pandemic. Table 4 presents the common psychiatric symptoms caused by COVID-19. Appendix D and E present a detailed statistics table of identified UMLS and the corresponding merged symptoms. Appendix Figure D.1 and Figure E.1 visually depict the distribution of most discussed physical and mental symptoms, respectively.

In terms of overall alignment with existing COVID-19 symptomatology studies,[19,20] our findings corroborate the presence of these symptoms. We identified all the symptoms of concern in the WHO report,[20] and almost all symptoms were recognized more frequently than in prior research using pre-defined lexicon.[19] According to the literature, shortness of breath is the most reported COVID-19 symptom, with other highly frequent symptoms including pain, cough, fatigue, fever, headache, anaphylaxis, asthma, taste altered, chill, respiratory symptoms, altered smell, and syncope. However, our list of symptoms exhibits a different distribution, as evidenced by several notable discrepancies in symptom frequencies across the three sources in our results. For instance, shortness of breath is reported at 9.4% in our approach but substantially higher at 18.6% in the WHO report, potentially reflecting variations in how dyspnea is recorded in hospitalized cases versus a broader dataset. Similarly, fever, consistently a top symptom, displays significant differences, with our approach reporting it at 7.2%, contrasting with the WHO Report at 87.9%, possibly due to differences in fever threshold for reporting in hospitalized cases. Notably, we identified a few symptoms that later proved to be typical symptoms of COVID-19 but were not recognized in early reports, including altered taste, altered smell, and alopecia. In addition, we also identified symptoms that were relatively infrequently discussed in prior research, including syncope, insomnia, and weight loss.

Moreover, our approach revealed additional symptoms related to mental health, which were challenging to capture through conventional human annotations and therefore overlooked by previous studies. The most common mental symptoms were anxiety, depression, distress, stress, and feeling bad.

Their overall frequency is also relatively high, reflecting the general increased psychological burden of people during the pandemic.[2]

**Table 3.** Occurrences and frequencies of the common symptoms identified by our dictionary, a pre-defined lexicon,[19] and World Health Organization (WHO) reporting.[20]

| Symptoms | Our dictionary N= 2,761,058 | Pre-defined lexicon N=948,478 | WHO Report N=55,924 |
|---|---|---|---|
| **Shortness of breath** | 259323 (9.4%) | 65949 (7.0%) | 10402 (18.6%) |
| **Pain** | 250265 (9.1%)[a] | 37814 (4.0%) | 8277 (14.8%)[b] |
| **Cough** | 199486 (7.2%) | 183039 (19.3%) | 37861 (67.7%) |
| **Fatigue** | 162768 (5.9%) | 147868 (15.6%) | 21307 (38.1%) |
| **Fever** | 155512 (5.6%) | 142752 (15.1%) | 49157 (87.9%) |
| **Headache** | 108398 (3.9%) | 101055 (10.7%) | 7606 (13.6%) |
| **Anaphylaxis** | 63798 (2.3%) | N/A[c] | N/A |
| **Asthma** | 63421 (2.3%) | N/A | N/A |
| **Taste altered** | 60753 (2.2%) | 38607 (4.1%) | N/A |
| **Chill** | 55164 (2.0%) | 27399 (2.9%) | 6375 (11.4%) |
| **Respiratory symptom** | 52532 (1.9%)[d] | N/A | N/A |
| **Smell altered** | 41480 (1.5%) | 26124 (2.8%) | N/A |
| **Syncope** | 40131 (1.5%) | N/A | N/A |
| **Discomfort** | 33423 (1.2%) | N/A | N/A |
| **Shock** | 30882 (1.1%) | N/A | N/A |
| **Nasal symptom** | 30446 (1.1%) | 7511 (0.8%) | 2684 (4.8%) |
| **Absence of sensation** | 29390 (1.1%) | N/A | N/A |
| **Nausea and/or Vomiting** | 28429 (1.0%) | 44429 (4.7 %) | 2796 (5%) |
| **Throat symptom** | 22180 (0.8%) | 43463 (4.6%) | 7773 (13.9%) |
| **Tremor** | 21699 (0.8%) | N/A | N/A |
| **Chest discomfort** | 21260 (0.8%) | 13382 (1.4%) | N/A |
| **Insomnia** | 19545 (0.7%) | N/A | N/A |
| **Sneezing** | 15778 (0.6%) | 37992 (4.0%) | N/A |
| **Coma** | 13843 (0.5%) | 13532 (1.4%) | N/A |
| **Influenza-like symptoms** | 13674 (0.5%) | N/A | N/A |
| **Weight loss** | 13307 (0.5%) | N/A | N/A |
| **Swelling** | 12109 (0.4%) | N/A | N/A |
| **Dizziness** | 11867 (0.4%) | 16628 (1.8%) | N/A |
| **Alopecia** | 11666 (0.4%) | N/A | N/A |
| **Diarrhea** | 7129 (0.3%) | N/A | 2069 (3.7%) |
| **Conjunctival congestion** | 1665 (0.1%) | N/A | 447 (0.8%) |

[a] Including myalgia (15620, 0.57%), arthralgia (8029, 0.29%), and other pains (226616, 8.22%)
[b] Specifically myalgia or arthralgia (8277, 14.8%)
[c] N/A: not applicable
[d] Including acute respiratory distress (26783, 0.97%), respiratory symptom (20321, 0.74%), decreased respiratory (2895, 0.1%), respiratory failure (2533, 0.09%)

**Table 4.** Occurrences and frequencies of the common mental symptoms identified by our dictionary.

| Symptoms | Count (frequency, %) | Symptoms | Count (frequency, %) |
|---|---|---|---|
| **Anxiety** | 241506 (8.7%) | **Posttraumatic stress disorder** | 17339 (0.6%) |
| **Depression** | 194278 (7.0%) | **Delusions** | 17004 (0.6%) |
| **Suffering** | 170345 (6.2%) | **Confusional state** | 14618 (0.5%) |
| **Stress** | 131780 (4.8%) | **Psychiatric symptom** | 8083 (0.3%) |
| **Feeling bad** | 91367 (3.3%) | **Suicidal** | 7987 (0.3%) |
| **Panic** | 90380 (3.3%) | **Moody** | 5076 (0.2%) |
| **Emotional upset** | 76660 (2.8%) | **Neurological symptom** | 4398 (0.2%) |
| **Anger reaction** | 50460 (1.8%) | **Loss of interest** | 3810 (0.1%) |
| **Paranoid disorder** | 41054 (1.5%) | **Flashbacks** | 3290 (0.1%) |
| **Feeling lonely** | 36594 (1.3%) | **Manic mood** | 2429 (0.1%) |

## 5. DISCUSSION

In public health research utilizing social media platforms, the selection of keywords is critical for identifying pertinent data. Yet, there remains a significant gap in standardizing this process, especially given the prevalence of colloquial language on such platforms. To address this gap, our study introduces a structured framework for creating comprehensive dictionaries tailored for public health research on social media. The constructed dictionary with COVID-19 symptom tweets, hierarchically arranged and integrated with the Unified Medical Language System (UMLS), presents a nuanced approach to capturing medical terminology, including its colloquial variants. This UMLS-aligned lexicon holds the potential to mitigate selection biases that might arise from incomplete keyword lists, thus enhancing the reliability and validity of research outcomes in this domain.

Leveraging NER, our methodology achieved automated and precise symptom identification, simplifying the task of extracting relevant data. With our normalization module designed for efficient candidate reduction and a mapping module that incorporates an iterative, interactive, and human-centric approach, we enhance the accuracy and efficiency of the dictionary, boosting its real-world applicability. The final dictionary allows for easy comprehension, editing, and extension. It is designed to be adept at handling vast datasets efficiently.

In contrasting our approach with traditional methods reliant on predefined lexicons,[19] our framework addresses critical limitations by dynamically mapping colloquial expressions to medical concepts. The observed disparities in symptom frequencies between our approach and other sources, such as WHO symptom reports, reflects the size and diversity of our study population. By using data from 948,478 cases, we include a wide range of disease severities and contexts. This extensive population differs from the hospitalized, severe cases emphasized in the WHO report. The substantial differences in symptom prevalence, such as the lower occurrence of fever and dyspnea in our approach, may similarly reflect the broader spectrum of COVID-19 manifestations in the general, outpatient population with more mild disease. Therefore, our results capture a comprehensive picture of the disease's clinical presentation.

Furthermore, our method provides distinct benefits compared to end-to-end deep learning models that classify post relevance in a binary fashion.[9,21] While these models are effective with a smaller volume of data, their reliance on extensive, labeled training datasets becomes a significant limitation. More crucially, as the volume of data grows, the feasibility of using deep learning models diminishes. This highlights the ongoing necessity for keyword identification, even in an era dominated by advanced large language models. Despite the near-perfect accuracy of these models in discerning relevant information,[22] deploying them on large-scale datasets incurs prohibitively high computational costs, financial burdens, and environmental impacts.[23–25] Notably, we found that CODER++, a popular UMLS named entity normalization model, had a low accuracy (~30%) in the context of social media data. As a result, extensive manual review was necessary to achieve target accuracy in the entity mapping process. This finding higlights limitations of existing models for entity map, and the need for models that can consistently comprehend the nuances of colloquial language in the continually evolving realm of social media.

While robust, our proposed framework has limitations. Firstly, the human-in-the-loop methodology adopted in the mapping module to improve model performance could introduce challenges related to scalability and timeliness, particularly when processing vast amounts of data. Additionally, the effectiveness of components like NER hinges upon the quality and diversity of the training data. Potential biases or overlooked colloquialisms in the training data could be mirrored in the framework's outcomes. Moreover, an inherent challenge not fully addressed in our approach is the complexity of mapping colloquialisms directly to UMLS concepts. As highlighted in previous literature,[17,19] certain phrases present inherent difficulties. For instance, the word "cough" in everyday language might ambiguously refer to both dry and wet coughs, whereas medical terminologies, such as those in the WHO report, differentiate between those two concepts. Although we have leveraged established lexicons like PASCLex to mitigate some of these discrepancies by adopting broader concepts, the inherent challenges of such mappings persist.

# 6. CONCLUSION

This study presents an innovative streamlined framework for the standardization and simplification of keyword selection processes in public health research on social media platforms. The resulted dictionary is publicly available and has shown improved accuracy and efficiency in symptom identification compared to traditional methods. Overall, this work contributes significantly to the computational public health field, providing a scalable, efficient tool for future epidemiological research on social media platforms.

# FUNDING

This research received no specific grant from any funding agency in public, commercial or not-for-profit sectors.

# AUTHOR CONTRIBUTION

Study design: Y.H. and J.Y.

Experiment implementation: Y.H and J.W.

Dictionary validation: S.L., M.L., Li, Y.Z., and D.F

Result interpretation: Y.H. and S.W.

Manuscript draft-up: Y.H.

Supervision: Y.J. and L.Z.

Manuscript revision and proofreading: All authors.

Y.H. takes responsibility for the integrity of this study.

# CONFLICT OF INTEREST STATEMENT

The authors have no conflicts of interest to declare relevant to the content of this article.

# ACKNOWLEDGEMENT

We thank Mr. Zifan Lin (Massachusetts Institute of Technology) for his contributions in proofreading the manuscript along the development of the study and assistance in the creation of figures.

# DATA AVAILABILITY STATEMENT

The dictionary developed in this study can be accessed at https://github.com/ningkko/COVID-drug. As Twitter (now renamed to X) does not permit the distribution of their data, we will furnish the tweet IDs used in this study upon request.

# REFERENCES

1. Institute of Medicine (US) Committee for the Study of the Future of Public Health. Washington (DC). Public Health as a Problem-Solving Activity: Barriers to Effective Action. In: *The Future of Public Health*. National Academies Press (US); 1988. Accessed January 21, 2024. https://www.ncbi.nlm.nih.gov/books/NBK218227/
2. Li M, Hua Y, Liao Y, et al. Tracking the Impact of COVID-19 and Lockdown Policies on Public Mental Health Using Social Media: Infoveillance Study. *J Med Internet Res*. 2022;24(10):e39676. doi:10.2196/39676
3. Hua Y, Jiang H, Lin S, et al. Using Twitter Data to Understand Public Perceptions of Approved versus Off-label Use for COVID-19-related Medications. *J Am Med Inform Assoc*. Published online July 18, 2022:ocac114. doi:10.1093/jamia/ocac114
4. Madhav N, Oppenheim B, Gallivan M, Mulembakani P, Rubin E, Wolfe N. Pandemics: Risks, Impacts, and Mitigation. In: Jamison DT, Gelband H, Horton S, et al., eds. *Disease Control Priorities: Improving Health and Reducing Poverty*. 3rd ed. The International Bank for Reconstruction and Development / The World Bank; 2017. Accessed January 21, 2024. http://www.ncbi.nlm.nih.gov/books/NBK525302/
5. Tsao SF, Chen H, Tisseverasinghe T, Yang Y, Li L, Butt ZA. What social media told us in the time of COVID-19: a scoping review. *Lancet Digit Health*. 2021;3(3):e175-e194. doi:10.1016/S2589-7500(20)30315-0
6. Rains SA. Big Data, Computational Social Science, and Health Communication: A Review and Agenda for Advancing Theory. *Health Commun*. 2020;35(1):26-34. doi:10.1080/10410236.2018.1536955
7. Lazer DMJ, Pentland A, Watts DJ, et al. Computational social science: Obstacles and opportunities. *Science*. 2020;369(6507):1060-1062. doi:10.1126/science.aaz8170
8. Vohra I, Nigam MS, Sakaria A, Kudari A, Rangaswamy N. Is Twitter Enough? Investigating Situational Awareness in Social and Print Media during the Second COVID-19 Wave in India. Published online November 29, 2022. doi:10.48550/arXiv.2211.16360

9.	Wu J, Wu X, Hua Y, Lin S, Zheng Y, Yang J. Exploring Social Media for Early Detection of Depression in COVID-19 Patients. *Proc ACM Web Conf 2023*. Published online April 30, 2023:3968-3977. doi:10.1145/3543507.3583867

10.	Bodenreider O. The Unified Medical Language System (UMLS): integrating biomedical terminology. *Nucleic Acids Res*. 2004;32(Database issue):D267-D270. doi:10.1093/nar/gkh061

11.	Lopez CE, Gallemore C. An augmented multilingual Twitter dataset for studying the COVID-19 infodemic. *Soc Netw Anal Min*. 2021;11(1):102. doi:10.1007/s13278-021-00825-0

12.	Devlin J, Chang MW, Lee K, Toutanova K. BERT: Pre-training of Deep Bidirectional Transformers for Language Understanding. *ArXiv181004805 Cs*. Published online May 24, 2019. Accessed April 19, 2022. http://arxiv.org/abs/1810.04805

13.	Zhou P, Wang Z, Chong D, et al. METS-CoV: A Dataset of Medical Entity and Targeted Sentiment on COVID-19 Related Tweets. In: *Proceedings of the 36th International Conference on Neural Information Processing Systems*. ; 2022. Accessed August 20, 2022. https://openreview.net/forum?id=GP1Ncd8nTgn

14.	Müller M, Salathé M, Kummervold PE. COVID-Twitter-BERT: A Natural Language Processing Model to Analyse COVID-19 Content on Twitter. Published online May 15, 2020. doi:10.48550/arXiv.2005.07503

15.	Jiang H, Hua Y, Beeferman D, Roy D. Annotating the Tweebank Corpus on Named Entity Recognition and Building NLP Models for Social Media Analysis. In: *Proceedings of the Thirteenth Language Resources and Evaluation Conference*. European Language Resources Association; 2022:7199-7208. Accessed October 25, 2022. https://aclanthology.org/2022.lrec-1.780

16.	Yuan Z, Zhao Z, Sun H, Li J, Wang F, Yu S. CODER: Knowledge-infused cross-lingual medical term embedding for term normalization. *J Biomed Inform*. 2022;126:103983. doi:10.1016/j.jbi.2021.103983

17.	Wang L, Foer D, MacPhaul E, Lo YC, Bates DW, Zhou L. PASCLex: A comprehensive post-acute sequelae of COVID-19 (PASC) symptom lexicon derived from electronic health record clinical notes. *J Biomed Inform*. 2022;125:103951. doi:10.1016/j.jbi.2021.103951

18.	Reimers N, Gurevych I. Sentence-BERT: Sentence Embeddings using Siamese BERT-Networks. *ArXiv190810084 Cs*. Published online August 27, 2019. Accessed April 30, 2022. http://arxiv.org/abs/1908.10084

19.	Wu J, Wang L, Hua Y, et al. Trend and Co-occurrence Network of COVID-19 Symptoms From Large-Scale Social Media Data: Infoveillance Study. *J Med Internet Res*. 2023;25(1):e45419. doi:10.2196/45419

20.	Report of the WHO-China Joint Mission on Coronavirus Disease 2019 (COVID-19). Accessed January 21, 2024. https://www.who.int/publications-detail-redirect/report-of-the-who-china-joint-mission-on-coronavirus-disease-2019-(covid-19)

21.	Biggers FB, Mohanty SD, Manda P. A deep semantic matching approach for identifying relevant messages for social media analysis. *Sci Rep*. 2023;13(1):12005. doi:10.1038/s41598-023-38761-y

22.	Zhu Y, Yuan H, Wang S, et al. Large Language Models for Information Retrieval: A Survey. Published online January 19, 2024. doi:10.48550/arXiv.2308.07107

23.	Zeng Q, Garay L, Zhou P, et al. GreenPLM: Cross-Lingual Transfer of Monolingual Pre-Trained Language Models at Almost No Cost. Published online May 26, 2023. doi:10.48550/arXiv.2211.06993

24.	Zhou H, Liu F, Gu B, et al. A Survey of Large Language Models in Medicine: Principles, Applications, and Challenges. Published online December 11, 2023. doi:10.48550/arXiv.2311.05112

25.	Hua Y, Liu F, Yang K, et al. Large Language Models in Mental Health Care: a Scoping Review. Published online January 1, 2024. doi:10.48550/arXiv.2401.02984

# APPENDICES

## Appendix A. Structure of the Named Entity Recognition Model

**Figure A.1** illustrates the structure of the study's NER model, providing insights into its functionality. In the figure, 'E' signifies an input embedding for a token, 'Trm' represents a transformer encoder, 'T' denotes the final hidden state, and 'C' represents the sentence embedding, although it was not utilized in our specific task.

The process begins with the tokenizer, which breaks down each tweet into individual tokens. Subsequently, the embedder maps these tokens onto a semantic space, enabling the model to comprehend their contextual meaning.

Towards the end of the model, we incorporated a linear layer and a SoftMax function. These additions facilitate the prediction of NER spans and labels pertaining to the mapped tokens. By leveraging this mechanism, the model can identify and classify the relevant entities within the tweet text.

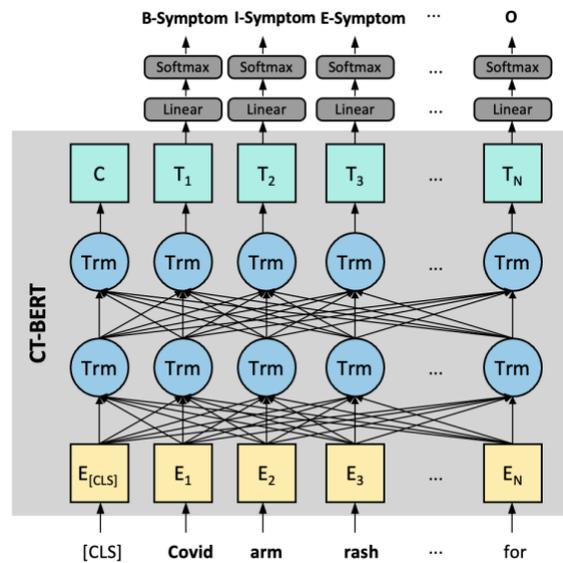

**Figure A.1** The structure of the CT-BERT-based NER model.

# Appendix B. Pre-cleaning the tweets

**Table B. 1.** Pre-cleaning patterns. During the matching process, each symptom entity and each keyword were padded with leading and trailing whitespaces to prevent mismatching.

| Phrases | Target |
|---|---|
| ’ | ' |
| could | can |
| can barely, can not, cant, could barely, hard to, harder to, having trouble, inability to, lost ability to, lost sense of, no one can, nobody can, not able to, not being able to, not to be able to, problem in, unable to | can't |
| issues, issue | problem |
| problem with, related problem, struggle with, struggled with, suffer, suffer from | problem |
| having no, having none of, lack of, little, dropped, low levels of, loss of, drop in, dropped | loss of |
| \b/\b | or |
| ° | degree |
| 'm, 've, am, are, been, being, being, going to, had , has, have, is, was, were | |
| he, her, him, his, I, it, It's, its, my, she, the, them, they, us, we, you, your | |
| all, even, likely, literally, many, massively, most, most of, mostly, much, probably, quite, really, rly, so, sometimes, super, very, pretty, hella, horribly, terribly | |

**Appendix C. Determining τ using similarity cut-offs.**

Figure C.1 shows the number of entities regarding changing thresholds. For fuzzy match, the elbow point **locates** at τ = 0.5. While for CODER++, the number of matched entities shows a sudden drop at τ = 0.8. Considering both methods, we set the threshold τ to be 0.8 given that the semantic part offers robustness to lexical variations, contextual understanding of nuanced terms, and a deeper grasp of complex medical jargon, making them more reliable for accurate entity-UMLS label mapping.

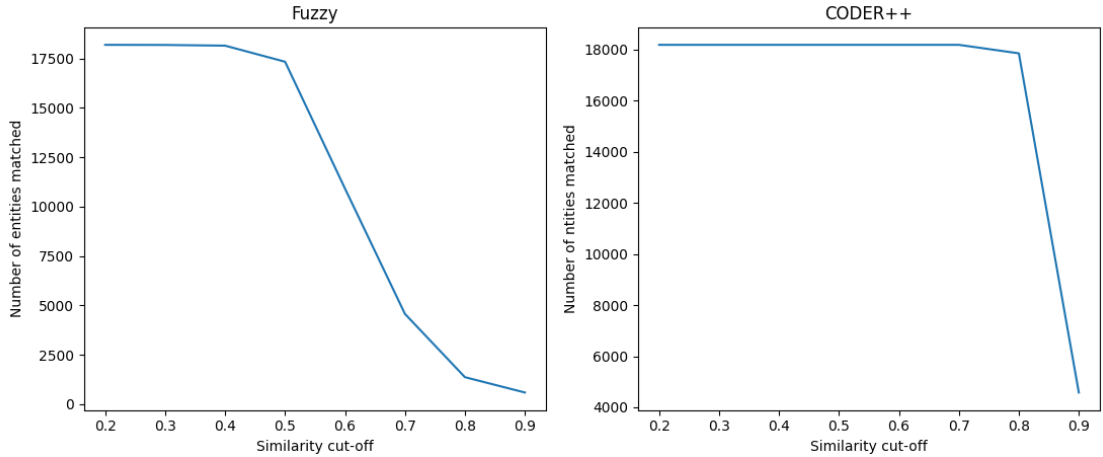

**Figure C.1.** Number of matches versus similarity cut-offs in the two normalization modules

**Appendix D. Frequency Statistics of Identified Physical Symptoms with Occurrences ≥ 500.**

**Table D.1 Detailed Frequency Statistics of Identified Physical Symptoms with Occurrences ≥ 500.**

| UMLS Concepts | Symptom name (merged) | Count (frequency, %) |
| --- | --- | --- |
| **Unable to breathe** | Shortness of breath | 225362 (8.16%) |
| **Cough** | Cough | 190143 (6.89%) |
| **Pain** | Pain | 165491 (5.99%) |
| **Fever** | Fever | 146651 (5.31%) |
| **Headache** | Headache | 92110 (3.34%) |
| **Fatigue** | Fatigue | 88609 (3.21%) |
| **Asthma** | Asthma | 63421 (2.3%) |
| **Anaphylaxis** | Anaphylaxis | 63260 (2.29%) |
| **Chill** | Chill | 54463 (1.97%) |
| **Tired** | Fatigue | 49426 (1.79%) |
| **Loss of taste** | Taste altered | 45828 (1.66%) |
| **Discomfort** | Discomfort | 33423 (1.21%) |
| **Shock** | Shock | 30882 (1.12%) |
| **Absence of sensation** | Absence of sensation | 29390 (1.06%) |
| **Loss of sense of smell** | Smell altered | 26304 (0.95%) |
| **Acute respiratory distress** | Respiratory symptom | 25395 (0.92%) |
| **Aching pain** | Pain | 23971 (0.87%) |
| **Collapse** | Syncope | 21926 (0.79%) |
| **Tremor** | Tremor | 21699 (0.79%) |
| **Respiratory symptom** | Respiratory symptom | 20321 (0.74%) |
| **Nausea** | Nausea and/or Vomiting | 18361 (0.66%) |
| **Generalized aches and pains** | Pain | 17553 (0.64%) |
| **Throat symptom** | Throat symptom | 17122 (0.62%) |
| **Blackout** | Syncope | 15787 (0.57%) |
| **Sneezing** | Sneezing | 15778 (0.57%) |
| **Exhaustion** | Fatigue | 15540 (0.56%) |
| **Dyspnea** | Shortness of breath | 15141 (0.55%) |
| **Coma** | Coma | 13843 (0.5%) |
| **Influenza-like symptoms** | Influenza-like symptoms | 13674 (0.5%) |
| **Recent weight loss** | Weight loss | 13307 (0.48%) |
| **Congestion** | Nasal symptom | 12748 (0.46%) |
| **Migraine** | Headache | 12743 (0.46%) |
| **Swelling** | Swelling | 12109 (0.44%) |
| **Alopecia** | Alopecia | 11666 (0.42%) |
| **Paralysis** | Paralysis | 11142 (0.4%) |
| **Chest pain** | Chest discomfort | 11100 (0.4%) |
| **Weakness** | Weakness | 10748 (0.39%) |
| **Somnolence** | Somnolence | 10209 (0.37%) |
| **Hemorrhage** | Hemorrhage | 10030 (0.36%) |
| **Abnormal breathing** | Shortness of breath | 9887 (0.36%) |
| **Problem with smell or taste** | Taste altered | 8349 (0.3%) |
| **Problem with smell or taste** | Smell altered | 8349 (0.3%) |
| **Cannot sleep at all** | Insomnia | 8124 (0.29%) |
| **Insomnia** | Insomnia | 7919 (0.29%) |
| **Sweating** | Sweating | 7489 (0.27%) |
| **Diarrhea** | Diarrhea | 7129 (0.26%) |
| **Unable to walk** | Unable to walk | 6896 (0.25%) |
| **Seizure** | Epilepsy | 6884 (0.25%) |
| **Nasal congestion** | Nasal symptom | 6629 (0.24%) |
| **Muscle pain** | Pain | 6520 (0.24%) |
| **Vomiting** | Nausea and/or Vomiting | 6504 (0.24%) |
| **Sight deteriorating** | Sight deteriorating | 6258 (0.23%) |
| **Soreness** | Soreness | 6006 (0.22%) |
| **Dry cough** | Cough | 5933 (0.21%) |
| **Delirium** | Delirium | 5843 (0.21%) |
| **Weight gain** | Weight gain | 5735 (0.21%) |
| **Chest discomfort** | Chest discomfort | 5586 (0.2%) |
| **Frailty** | Frailty | 5579 (0.2%) |
| **Cramp** | Cramp | 5569 (0.2%) |
| **Myalgia and myositis unspecified** | Pain | 4855 (0.18%) |
| **Tight chest** | Chest discomfort | 4574 (0.17%) |
| **Chronic fatigue syndrome** | Fatigue | 4482 (0.16%) |
| **Unable to stand** | Unable to stand | 4480 (0.16%) |
| **Feeling feverish** | Fever | 4409 (0.16%) |
| **Dizziness** | Dizziness | 4370 (0.16%) |
| **Choking** | Choking | 4365 (0.16%) |
| **Fibromyalgia** | Pain | 4245 (0.15%) |
| **Taste sense altered** | Taste altered | 4230 (0.15%) |
| **Joint pain** | Pain | 4219 (0.15%) |
| **Throat irritation** | Throat symptom | 4044 (0.15%) |
| **Earache symptoms** | Earache symptoms | 3881 (0.14%) |
| **Tinnitus** | Tinnitus | 3860 (0.14%) |
| **Lightheadedness** | Dizziness | 3789 (0.14%) |
| **Erectile dysfunction** | Erectile dysfunction | 3765 (0.14%) |
| **Watery eye** | Watery eye | 3756 (0.14%) |
| **Chronic pain** | Pain | 3526 (0.13%) |
| **Winded** | Shortness of breath | 3433 (0.12%) |
| **Palpitations** | Palpitations | 3396 (0.12%) |
| **Backache** | Pain | 3239 (0.12%) |
| **Fever symptoms** | Fever | 3217 (0.12%) |
| **Stomach ache** | Stomach discomfort | 3200 (0.12%) |
| **Loss of appetite** | Appetite problem | 3157 (0.11%) |
| **Bleeding from nose** | Nasal symptom | 3154 (0.11%) |
| **Vertigo** | Dizziness | 3089 (0.11%) |
| **Wheezing** | Shortness of breath | 3040 (0.11%) |
| **Acneiform eruption** | Acneiform eruption | 3028 (0.11%) |
| **Nasal symptom** | Nasal symptom | 3000 (0.11%) |
| **Decreased respiratory function** | Respiratory symptom | 2895 (0.1%) |
| **Blindness** | Blindness | 2836 (0.1%) |
| **Sense of smell altered** | Smell altered | 2830 (0.1%) |
| **Pain in upper limb** | Pain | 2801 (0.1%) |
| **Unconscious** | Unconscious | 2676 (0.1%) |
| **Malaise** | Fatigue | 2605 (0.09%) |
| **Apnea** | Apnea | 2587 (0.09%) |
| **Respiratory failure** | Respiratory symptom | 2533 (0.09%) |
| **Tachycardia** | Tachycardia | 2517 (0.09%) |
| **Appetite problem** | Appetite problem | 2503 (0.09%) |
| **Gagging** | Nausea and/or Vomiting | 2494 (0.09%) |
| **Sinus pain** | Nasal symptom | 2355 (0.09%) |
| **Wheal** | Wheal | 2342 (0.08%) |
| **Loss of memory** | Memory dysfunction | 2341 (0.08%) |
| **Undernourished** | Undernourished | 2228 (0.08%) |
| **Unusual smell in nose** | Smell altered | 2196 (0.08%) |
| **Difficulty sleeping** | Insomnia | 2157 (0.08%) |
| **Has tingling sensation** | Tingling sensation | 2146 (0.08%) |
| **Lethargy** | Fatigue | 2106 (0.08%) |
| **Aching headache** | Headache | 2027 (0.07%) |
| **Toothache** | Toothache | 2028 (0.07%) |
| **Unable to move** | Unable to move | 2029 (0.07%) |
| **Shoulder pain** | Pain | 2031 (0.07%) |
| **Myopia** | Myopia | 2032 (0.07%) |
| **Deafness** | Deafness | 2034 (0.07%) |
| **Sense of smell impaired** | Smell altered | 2035 (0.07%) |
| **Activity intolerance** | Activity intolerance | 2037 (0.07%) |
| **Reflux** | Reflux | 1772 (0.06%) |
| **Hiccoughs** | Hiccoughs | 1763 (0.06%) |
| **Upset stomach** | Stomach discomfort | 1762 (0.06%) |
| **Impaired cognition** | Impaired cognition | 1701 (0.06%) |
| **Ulcer** | Ulcer | 1615 (0.06%) |
| **Eczema** | Eczema | 1527 (0.06%) |
| **Numbness** | Numbness | 1511 (0.05%) |
| **Hearing loss** | Hearing loss | 1498 (0.05%) |
| **Night sweats** | Sweating | 1457 (0.05%) |
| **Syncope** | Syncope | 1441 (0.05%) |
| **Wound pain** | Pain | 1390 (0.05%) |
| **Respiratory distress** | Respiratory symptom | 1388 (0.05%) |
| **Persistent cough** | Cough | 1372 (0.05%) |
| **Neuralgia** | Neuralgia | 1366 (0.05%) |
| **Heartburn** | Heartburn | 1338 (0.05%) |
| **Lymphadenopathy** | Lymphadenopathy | 1299 (0.05%) |
| **Heavy pain** | Pain | 1261 (0.05%) |
| **Acid reflux** | Acid reflux | 1247 (0.05%) |
| **Gastrointestinal symptom** | Gastrointestinal symptom | 1242 (0.04%) |
| **Dissociation** | Dissociation | 1237 (0.04%) |
| **Feels hot/feverish** | Fever | 1235 (0.04%) |
| **Spasm** | Cramp | 1211 (0.04%) |
| **Cold sweat** | Sweating | 1209 (0.04%) |
| **Unpleasant taste in mouth** | Taste altered | 1201 (0.04%) |
| **Amnesia** | Memory dysfunction | 1153 (0.04%) |
| **Hip pain** | Pain | 1115 (0.04%) |
| **Blood coagulation disorder** | Blood coagulation disorder | 1110 (0.04%) |
| **Cramping** | Cramp | 1097 (0.04%) |
| **Flushing** | Flushing | 1095 (0.04%) |
| **Cognitive disorder** | Impaired cognition | 1091 (0.04%) |
| **Hoarse** | Hoarse | 1090 (0.04%) |
| **Severe pain** | Pain | 1089 (0.04%) |
| **Nausea present** | Nausea and/or Vomiting | 1070 (0.04%) |
| **Forgetful** | Memory dysfunction | 1058 (0.04%) |
| **Posterior rhinorrhea** | Nasal symptom | 1043 (0.04%) |
| **Hypothermia** | Hypothermia | 1026 (0.04%) |
| **Pain in throat** | Throat symptom | 1014 (0.04%) |
| **Knee pain** | Pain | 989 (0.04%) |
| **Heat syncope** | Syncope | 977 (0.04%) |
| **Red eye** | Conjunctival congestion | 968 (0.04%) |
| **Neck pain** | Pain | 967 (0.04%) |
| **Unable to concentrate** | Unable to concentrate | 948 (0.03%) |
| **Tunnel visual field constriction** | Tunnel visual field constriction | 939 (0.03%) |
| **Weight change** | Weight change | 932 (0.03%) |
| **Psoriasiform rash** | Psoriasiform rash | 926 (0.03%) |
| **Increasing breathlessness** | Shortness of breath | 910 (0.03%) |
| **Constant pain** | Pain | 901 (0.03%) |
| **Memory dysfunction** | Memory dysfunction | 897 (0.03%) |
| **Sinus headache** | Headache | 882 (0.03%) |
| **Eruption** | Eruption | 879 (0.03%) |
| **Hacking cough** | Cough | 879 (0.03%) |
| **Abdominal pain** | Pain | 878 (0.03%) |
| **Pain in lower limb** | Pain | 874 (0.03%) |
| **Blister** | Blister | 872 (0.03%) |
| **Disorientated** | Disorientated | 871 (0.03%) |
| **Pustule** | Pustule | 867 (0.03%) |
| **Ankle pain** | Pain | 864 (0.03%) |
| **Excessively deep breathing** | Shortness of breath | 843 (0.03%) |
| **Cardiac arrhythmia** | Cardiac arrhythmia | 828 (0.03%) |
| **Congestion of nasal sinus** | Nasal symptom | 818 (0.03%) |
| **Excruciating pain** | Pain | 787 (0.03%) |
| **Difficulty walking** | Difficulty walking | 777 (0.03%) |
| **Low back pain** | Pain | 773 (0.03%) |
| **Sleep deprivation** | Insomnia | 713 (0.03%) |
| **Labored breathing** | Shortness of breath | 707 (0.03%) |
| **Stomach cramps** | Stomach discomfort | 702 (0.03%) |
| **Feels cold** | Chill | 701 (0.03%) |
| **Rhinitis** | Nasal symptom | 699 (0.03%) |
| **Bleeding eye** | Conjunctival congestion | 697 (0.03%) |
| **Sensation of hot and cold** | Sensation of hot and cold | 676 (0.02%) |
| **Itching of eye** | Itching of eye | 664 (0.02%) |
| **Snoring** | Snoring | 656 (0.02%) |
| **Sciatica** | Sciatica | 648 (0.02%) |
| **Chronic headache disorder** | Headache | 644 (0.02%) |
| **Not getting enough sleep** | Insomnia | 632 (0.02%) |
| **Sore on skin** | Sore on skin | 631 (0.02%) |
| **Dizzy spells** | Dizziness | 619 (0.02%) |
| **Visual impairment** | Visual impairment | 614 (0.02%) |
| **Productive cough** | Cough | 596 (0.02%) |
| **Metallic taste** | Taste altered | 591 (0.02%) |
| **Pins and needles** | Tingling sensation | 585 (0.02%) |
| **Sleep walking disorder** | Sleep walking disorder | 580 (0.02%) |
| **Burning sensation** | Burning sensation | 578 (0.02%) |
| **Chronic cough** | Cough | 563 (0.02%) |
| **Abnormal taste in mouth** | Taste altered | 554 (0.02%) |
| **Indigestion** | Indigestion | 542 (0.02%) |
| **Anaphylactic shock** | Anaphylaxis | 538 (0.02%) |
| **Edema** | Edema | 527 (0.02%) |
| **Pain in eye** | Pain in eye | 520 (0.02%) |
| **Poor balance** | Poor balance | 500 (0.02%) |

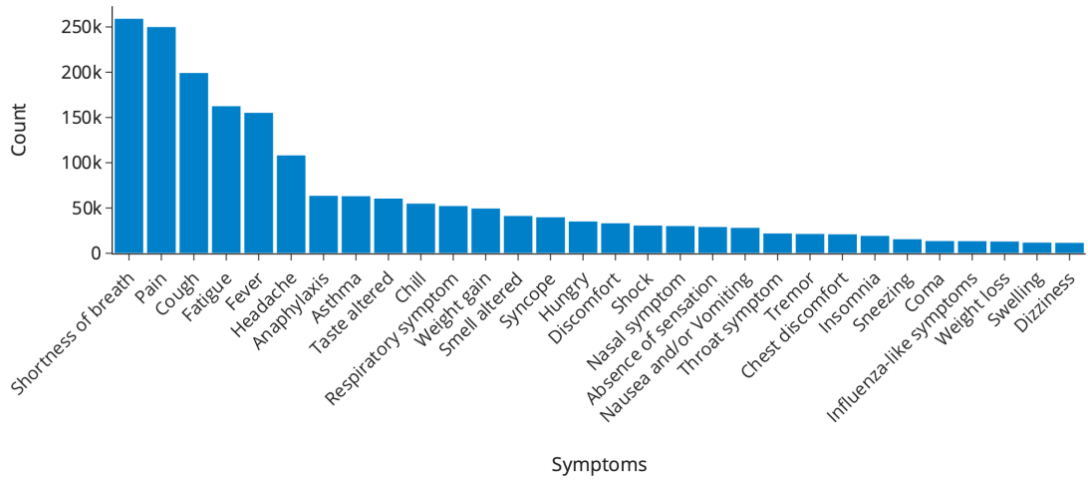

**Figure D.1.** Top 30 most discussed physical symptoms in COVID-19-related tweets.